\documentclass[preprint,12pt,authoryear]{elsarticle}
\usepackage{amsthm}
\usepackage{amsmath}
\usepackage{amssymb}
\usepackage{amsfonts}
\usepackage{amsthm}
\usepackage{bm}
\usepackage{latexsym}
\usepackage{graphicx}
\usepackage{setspace}
\usepackage{array}
\usepackage{float}
\usepackage{url}
\usepackage{multirow}
\usepackage{soul}
\usepackage{xcolor}
\usepackage{enumerate}
\usepackage[shortlabels]{enumitem}
\usepackage{algorithm}
\usepackage[noend]{algpseudocode}
\usepackage{afterpage}
\usepackage{lineno}
\usepackage[margin=1.25in]{geometry}
\usepackage{verbatim}
\usepackage{fancyvrb}
\usepackage{threeparttable}
\usepackage{setspace}

\numberwithin{remark}{section}
\begin{document}

\begin{frontmatter}

\title{Multi-rank sparse hierarchical clustering}
\author{Hongyang Zhang}
\author{Ruben H. Zamar}
\address{Department of Statistics, University of British Columbia, 3182-2207 Main Mall, Vancouver, British Columbia V6T 1Z4, Canada}

\begin{abstract}
{\small 
There has been a surge in the number of large and flat data sets  -- data sets containing a large number of features and a relatively small number of observations --  due to the growing ability to collect and store information in medical research and other fields. Hierarchical clustering is a widely used clustering tool.  In hierarchical clustering,  large and flat data sets may allow for a better coverage of clustering features (features that help explain the true underlying clusters) but, such data sets usually include a large fraction of noise features (non-clustering features) that may hide the underlying clusters. \citet{witten2010framework} proposed a sparse hierarchical clustering framework to cluster the observations using an adaptively chosen subset of the features, however, we show that this framework has some limitations when the data sets contain clustering features with complex structure. In this paper, we propose the Multi-rank sparse hierarchical clustering (MrSHC). We show that, using simulation studies and real data examples, MrSHC produces superior feature selection and clustering performance comparing to the classical (of-the-shelf) hierarchical clustering and the existing sparse hierarchical clustering framework.
}
\end{abstract}

\end{frontmatter}

\section{Introduction}
The performance of existing clustering algorithms  can be distorted when the number of variables is large and many of them contain no information about the cluster structure. Furthermore, interpretability can be impeded when the clustering procedure uses a large number of variables.  Thus, clustering algorithms that can simultaneously perform cluster analysis and feature selection are in demand.

Here we focus on hierarchical clustering, one of the most widely used clustering algorithms. Hierarchical clustering categorizes observations into a hierarchical set of groups organized in a tree structure called dendrogram. Hierarchical clustering has a broad range of applications such as microarray data analysis,  digital imaging, stock prediction, text mining, etc.

There are several proposals for feature selection for other clustering methods  such as K-means (e.g.\ \citet{witten2010framework}, \citet{sun2012regularized}) and model-based clustering (e.g.\ \citet{raftery2006variable}, \citet{pan2007penalized}, \citet{wang2008variable}, \citet{xie2008penalized}). However, there has been less research for the case  of hierarchical clustering. A brief survey of such proposals is given below. 

 Let $\bm{X}$ be an $n \times p$ data matrix, with $n$ observations and $p$ features. Let $d_{i, i'}=d(\bm{x}_i, \bm{x}_{i'})$ be a measure of dissimilarity between observations $\bm{x}_i$ and $\bm{x}_{i'}$ ($1\le i, i' \le n$), which are the rows $i$ and $i'$ of the data matrix $\bm{X}$. We will assume that $d$ is additive in the features: $d(\bm{x}_i , \bm{x}_{i'})=d_{i, i'}=\sum_{j=1}^p d_{i, i', j}$, where $d_{i, i', j}$ indicates the dissimilarity between observations $i$ and $i'$ along feature $j$. Unless specified otherwise, our examples and simulations take $d$ equal to the squared Euclidean distance, $d_{i,i',j} = (X_{ij}-X_{i'j})^2$. However, other dissimilarity measures are possible, such as the absolute difference $d_{i,i',j} = |X_{ij} - X_{i'j}|$.

\citet{friedman2004clustering} proposed \textit{clustering objects on subsets of attributes} (COSA). COSA employs a criterion, related to a weighted version of K-means clustering, to automatically detect  subgroups of objects that preferentially cluster on subsets of the attribute variables rather than on all of them simultaneously. An extension of COSA for hierarchical clustering was also proposed. The  algorithm is quite complex and requires multiple tuning parameters. Moreover, as noted by \citet{witten2010framework}, this proposal does not truly result in a sparse clustering because all the variables have nonzero weights.

 \citet{witten2010framework} proposed a new framework for sparse clustering that can be applied to procedures that optimize a criterion of the form 
\begin{equation}\label{eq:optim-general}
\max_{\Theta\in G}\left\{ \sum_{j=1}^p f_j(\mathbf{X}_j, \Theta) \right\},
\end{equation}
where
 $\mathbf{X}_j = (X_{1j},X_{2j},...,X_{nj})^T \in \mathbb R^n$ denotes the observed $j$-th feature, 
 each $f_j(\mathbf{X}_j, \Theta)$ is a function that solely depends on the $j$-th feature and $\Theta$ is a set of unknown parameters taking values on $G$.
To introduce sparsity \citet{witten2010framework}  modified criterium (\ref{eq:optim-general})  as follows:
%
\begin{equation}\label{eq:optim-general-sparse}
\max_{\bm{w}, \Theta\in G}\left\{ \sum_{j=1}^p w_j f_j(\mathbf{X}_j, \Theta) \right\} ~\textrm{subject to}~ ||\bm{w}||_2^2 \le 1, \, ||\bm{w}||_1 \le s \, \text{and } w_j \ge 0.
\end{equation}
Here $\bm{w}=(w_1, w_2, \cdots, w_p)$ is a vector of weights for each feature, $||\bm w||_2^2$ is squared L2-norm on $\bm w$, and $||\bm w||_1$ is L1-norm on $\bm w$. A feature with zero-weight is clearly not used in the criterion.

Hierarchical clustering 
does not optimize a criterium like (\ref{eq:optim-general})  and, therefore,  does not directly fit  into \citet{witten2010framework} sparse  clustering framework (\ref{eq:optim-general-sparse}). To overcome this difficulty they
casted the dissimilarity matrix $\left\{ d_{i, i'}\right\}_{n\times n}$ as the solution of an optimization problem as follows:
\begin{equation}\label{eq:optim-dissim}
\max_{\mathbf{U} \in \mathbb R^{n\times n}}\left\{\sum_{j=1}^p \sum_{i, i'=1}^n d_{i,i',j} U_{i, i'} \right\} ~ \textrm{subject to}~ \sum_{i,i'=1}^n U^2_{i,i'} \le 1.
\end{equation}
 It can be shown that the solution $\widehat{U}_{i,i'}$ to (\ref{eq:optim-dissim}) is proportional to the dissimilarity matrix, that is, $\widehat{U}_{i,i'} \propto  d_{i,i'}$. The criterion in  (\ref{eq:optim-dissim}) is a special case of (\ref{eq:optim-general}) when we let
$f_j(\mathbf{X}_j, \Theta)=\sum_{i, i'=1}^n d_{i,i',j} U_{i, i'}$. Now sparse hierarchical clustering can be achieved by obtaining a sparse dissimilarity matrix. Now
\textit{the sparse hierarchical clustering criterion} 
can be defined as follows:
\begin{equation}\label{eq:optim-dissim-sparse}
\max_{\bm{w},\mathbf{U} \in \mathbb R^{n\times n}}\left\{\sum_{j=1}^p w_j \sum_{i, i'=1}^n d_{i,i',j} U_{i, i'} \right\} ~\textrm{subject to}~ \sum_{i,i'=1}^n U^2_{i,i'} \le 1, ||\bm{w}||_2^2 \le 1, \, ||\bm{w}||_1 \le s.
\end{equation}
The constraint $w_j \ge 0$ has been removed because $d_{i,i',j} \ge 0$ for all $1 \le i, i' \le n$ and $1 \le j \le p$.
The solution to (\ref{eq:optim-dissim-sparse}) can be obtained using sparse principal component (SPC) proposed in \citet{witten2009penalized} as follows:
Let $\bm{u}$ be a vector of length $n^2$ that contains all elements in $(U_{i,i'})_{n \times n}$ and $\bm{D}$ be a $n^2 \times p$ matrix whose $j$-th column contains the $n^2$ elements of $\left\{ d_{i, i', j} \right\}_{n\times n}$ -- the dissimilarity matrix calculated from the $j$-th feature alone. Now the criterion in (4) is equivalent to the following:
\begin{equation}\label{eq:optim-SPC}
\max_{\bm{w}, \bm{u}}\left\{\bm{u}^T\bm{D}\bm{w}\right\} ~\textrm{subject to}~ ||\bm{u}||_2^2 \le 1, ||\bm{w}||_2^2 \le 1, ||\bm{w}||_1 \le s.
\end{equation}
This reduces to applying SPC on \emph{the transformed dissimilarity matrix}, $\bm{D}$. It can also be shown that the solution to (\ref{eq:optim-SPC}) satisfies: $\widehat{\bm{u}}^T \propto \bm{D}\widehat{\bm{w}}$. As $\widehat{\bm{w}}$ is sparse, so is $\widehat{\bm{u}}$. Thus, by re-arranging the elements in  $\widehat{\bm{u}}$ into a $n\times n$ dissimilarity matrix $\widehat{\bm{U}}$, we obtain a sparse dissimilarity matrix which only contains the information from a subset of selected features. Finally, sparse hierarchical clustering can be obtained by 
 applying classical hierarchical clustering on the sparse dissimilarity matrix $\widehat{\mathbf{U}}$.
\citet{witten2010framework} showed, using a simulated dataset and a genomic dataset, that their proposed sparse hierarchical clustering results in more accurate identification of the underlying clusters and more interpretable results than standard hierarchical clustering and COSA when applied on datasets with noise features.

However, as we show in the following sections, the SHC framework has its limitations, especially when the features contain complex structures. To remedy these limitations, we propose the Multi-rank Sparse Hierarchical Clustering (MrSHC) framework which proves to outperform the traditional hierarchical clustering and the SHC in both simulated and real data examples.

The rest of the paper is organized as follows. In Section~2, we list the limitations of the SHC framework with a motivating example. Section~3 presents the proposed Multi-rank Sparse Hierarchical Clustering (MrSHC) framework. We presents the results from simulation studies and real data examples in Section~4 and 5 respectively. Finally, in Section~6, we conclude with some remarks.

\section{Limitations of \citet{witten2010framework}'s sparse hierarchical clustering} \label{sec:limitation}

SHC essentially applies SPC criterion to a transformed dissimilarity matrix $\bm{D}^*$ and obtains the best rank-1 sparse approximation of $\bm{D}^*$ given a sparsity constraint, i.e.\ criterion (\ref{eq:optim-SPC}). The clustering features are chosen according to the non-zero loadings in the first sparse principal component resulted from the rank-1 sparse approximation. This approach share the same limitations when the clustering features may not be fully identified by a single sparse principal component. In other words, the clustering features may not be properly recovered by only rank-1 approximation. The following simulated example illustrates this situation.

We generate a data set $\bm{X}$ as follows:
$\bm{X}$ contains $n=20$ observations with $p=15$ features, i.e.\ $\bm{X}_{n\times p}=(\bm{x}_1, \bm{x}_2, \cdots, \bm{x}_{n})^T$, where $\bm{x}_i=(x_{i,1}, x_{i,2}, \cdots, x_{i,p})^T$, $1\le i \le n$. The observations are organized in four clusters of size 5. Let $Y_i$, ($i = 1, \cdots, n$) denote the cluster memberships. Then $x_{ij}$ ($i = 1, \cdots, n$) is generated from $N(\mu_j(Y_i), 0.1)$ for $j = 1, \cdots, 4$, and $N(\mu_j(Y_i), 1)$ for $j = 5, \cdots, p$.

A sketch of $\bm\mu_j(Y_i)$ is presented in the table below:
\begin{table}[h]
\centering
\begin{tabular}{c|cccc|c}
  \hline
  $Y_i$ & $\mu_1(Y_i)$ & $\mu_2(Y_i)$ & $\mu_3(Y_i)$ & $\mu_4(Y_i)$ & $\mu_5(Y_i) \cdots \mu_{P}(Y_i)$\\
    \hline
  1 & 1 & 1 & 1 & 1 & $0 ~~~ \cdots ~~~ 0$\\
  2 & -1 & -1 & 1 & 1 &  $0 ~~~ \cdots ~~~ 0$\\
  3 & -1 & -1 & -1 & -1 &$0 ~~~ \cdots ~~~ 0$ \\
  4 & 1 & 1 & -1 & -1 &  $0 ~~~ \cdots ~~~ 0$\\
  \hline 
\end{tabular}
\end{table}

\begin{figure}[htbp]
\begin{center}
  \includegraphics[scale = 0.25]{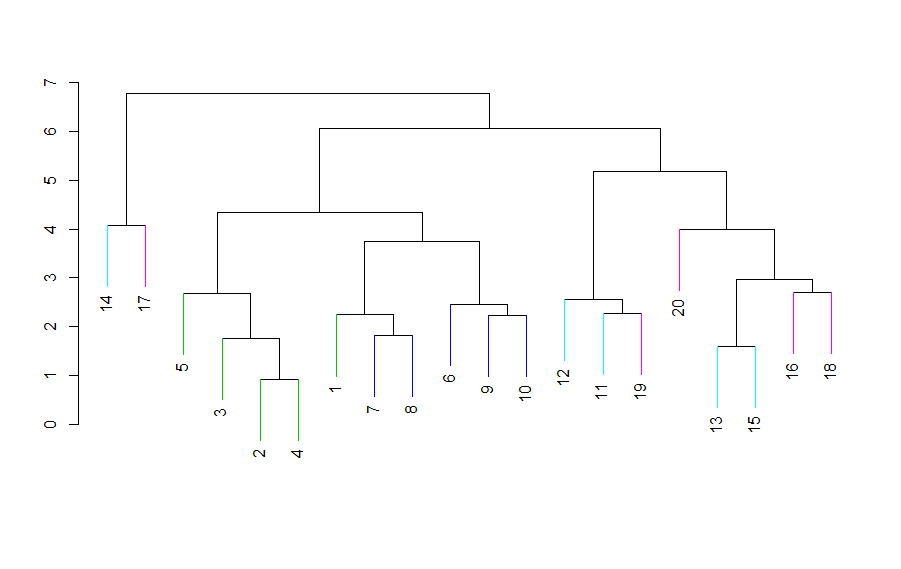}
  \end{center}
  \caption{Dendrogram generated with the first four chosen features $\{V_{13}, V_{11}, V_6,$ \textcolor[rgb]{0.98,0.00,0.00}{\textbf{$V_1$}}\} from Witten and Tibshirani's sparse hierarchical clustering framework}
  \label{exampleI}
\end{figure}

We apply SHC to $\bm{X}$. By gradually increasing the sparsity constraint, we obtain the sequence of the first 9 chosen variables $\{V_{13}, V_{11}, V_6$, \textcolor[rgb]{0.98,0.00,0.00}{\textbf{$V_1$}},  \textcolor[rgb]{0.98,0.00,0.00}{\textbf{$V_2$}},  $V_{14}$, $V_8$, \textcolor[rgb]{0.98,0.00,0.00}{\textbf{$V_4$}}, \textcolor[rgb]{0.98,0.00,0.00}{\textbf{$V_3$}}\}. The first three chosen features are noise features. As a result, the dendrogram generated from the first four chosen features (which is suggested by \citet{witten2010framework}'s auto-selection method) gives mixed clusters (See Figure~\ref{exampleI}). The clustering result is still unsatisfactory even if seven variables are chosen (results not show here). Moreover, five noise features are selected before all the four clustering features are chosen.

%

\section{Multi-rank sparse hierarchical clustering}
To remedy this limitation of SHC, we propose the multi-rank sparse hierarchical clustering (MrSHC). Similar to SHC, MrSHC uses SPC as an important building block for feature selection, but MrSHC is different in the following aspects. 
\begin{itemize}
\item MrSHC applies SPC directly to the original data $\bm{X}$. In comparison of applying SPC to a transformed dissimilarity matrix of size $n^2 \times p$, this approach is more computationally efficient, more intuitive and proves to be effective given MrSHC's superior performance in simulation studies and real data examples.
\item Re-assign weight 1 to all the features with non-zero weights in sparse PCs. We call the re-assigned weights ``indicator weights''. Indicator weights simply reflect whether features are selected, without further weighting on each feature, thus, facilitates the interpretation of the clustering results. Indicator weights also allow MrSHC to degenerate to classical hierarchical clustering if no sparsity constraint is applied. 
\item MrSHC identifies and recovers the clustering features using multi-rank sparse approximation through SPC. In other words, the clustering features are chosen according to the non-zero loadings in multiple sparse PCs. This allows MrSHC to adapt to features with more complex structures.
\end{itemize}

MrSHC is very different from the traditional approach where high-dimensional data are clustered based on the first few principal components. First, MrSHC applies SPC instead of traditional PCA to the data, also, MrSHC chooses the original features for clustering. The chosen features can be closely approximated by sparse low-rank approximations, in other words, they should have similar patterns of variations (e.g.\ if the features can be closely approximated by rank-1 approximation, then each of the features should have similar variation as the first PC). In clustering, similar patterns of variations usually represent information of clusters, and thus, the features chosen from MrSHC should contain key informations of clusters. This is confirmed by the simulated and real data examples in later sections.

We outline the MrSHC framework under the cases in Table~\ref{cases}.
\begin{table}[htbp]
 \centering
\begin{tabular}{c|c|c}
  \hline
Case & $q$ & $r$\\
\hline
1 & Known & Known\\
2 & Known & Unknown\\
3 & Unknown & Unknown\\
\hline
\end{tabular}
\caption{Different cases for MrSHC; $q$ is the target number of selected features and $r$ is the rank of the SPC approximation}\label{cases}.
\end{table}

\subsection{Case~1: Known $q$ and $r$}
Suppose the target number of clustering features $q$ and the appropriate rank $r$ of the SPC approximation are known. MrSHC applies SPC to $\bm{X}$ and obtain the first $r$ sparse PCs. In MrSHC, a feature is considered selected if it has a non-zero loading in any of the $r$ sparse PCs. With an appropriately chosen sparsity constraint $\lambda$, we can get $q$ or approximately $q \pm d$ (say $d = 1$) chosen features from the $r$ sparse PCs. MrSHC chooses $\lambda$ using a bi-section approach presented in Algorithm~\ref{alg:bisection}.

\begin{algorithm}[h!]
\caption{Feature set selection}\label{alg:bisection}
\begin{algorithmic}[1]
\\ \textbf{Input:} $\bm{X}_{n\times p}$, $q$, $r$, $d$ (default $d = 0$).
\\  Assign $\lambda^{-} = 1$ and $\lambda^{+} = \sqrt{n}$ ($\lambda^{+}$ can be set smaller in practice).
\\ \textbf{Repeat} Step 4-8.
\\ Apply SPC to $\bm{X}$ with $\lambda = (\lambda^{-}_k + \lambda^{+}_k)/2$; obtain the first $r$ sparse PCs.
\\ $q^* :=$ the number of variables with non-zero loadings in any of the $r$ sparse PCs.
\\ $\mathcal{C}_{r} := $ the set of $q^*$ chosen variables.
\\ \textbf{Break} if $q - d \le q^{*} \le q + d$.
\\ If $q^{*} > q + d$, $\lambda^{+} = \lambda^{*}$; if $q^{*} < q - d$, $\lambda^{-} = \lambda^{*}$.
\\ \textbf{Output: } $\mathcal{C}_{r}$, $q^*$.
\end{algorithmic}
\end{algorithm}

We have seen in the simulations and real data examples that Algorithm~\ref{alg:bisection} will finish in a few iterations.

Given $q$ and $r$, a set of chosen features $\mathcal{C}_r$ can be obtained from Algorithm~\ref{alg:bisection}. Then MrSHC simply generates a dendrogram (with any linkage of choice) based on the features in $\mathcal{C}_r$.

\subsection{Case~2: Known $q$, unknown $r$} \label{sec:+q-r}
Suppose $q$ is known, but not $r$. To choose $r$, MrSHC first applies Algorithm~\ref{alg:bisection} with increasing ranks $r_i$, $i = 1, \cdots, R$ (different $R$ can be chosen; we choose $R = 8$ here). Given rank $r_i$, let $\mathcal{C}_{r_i}$ denote the candidate feature set obtained from Algorithm~\ref{alg:bisection}. Different ranks $r_i$ will be compared through their corresponding $\mathcal{C}_{r_i}$. MrSHC assesses the quality of a feature set $\mathcal{C}_{r_i}$ through the dendrogram generated from its features. To be more specific, a feature set $\mathcal{C}_{r_i}$ is evaluated according to the following aspects.
\begin{itemize}
\item The number of well-separated clusters discovered from the dendrogram generated from $\mathcal{C}_{r_i}$. MrSHC uses a multi-layer pruning approach to obtain the well-separated clusters from a dendrogram. We introduce a ``reference number of clusters" in the multi-layer pruning to facilitate later comparisons (described below). 
\item The degree of separation of the discovered clusters, which is evaluated by silhouette values \citep{rousseeuw1987silhouettes}.
\end{itemize}
Given the number of discovered clusters and the silhouette values, an iterative selection approach is proposed to choose the final rank $r$. 

Multi-layer pruning (MLP) prunes the dendrogram from the top to the bottom, with each split evaluated by the Gap statistics \citep{tibshirani2001estimating}. We introduce a ``reference number of clusters'' $K$ in MLP, which is both an ``upper bound'' and a ``lower bound''. It is an upper bound of the number of clusters discovered in MLP. When $K$ is chosen properly, MLP will produce labels for $K$ clusters for most of the input dendrograms generated from $\mathcal{C}_{r_i}$, $i = 1, \cdots, R$. This facilitates later comparisons since labels with different number of clusters are in general difficult to compare. It is also a lower bound of the number of clusters that are expected to be discovered. If less than $K$ clusters are discovered from a dendrogram according to MLP, such a dendrogram and its corresponding feature set are considered to be of low quality since key clusters may be missing. Therefore, such feature sets and their corresponding ranks are screened out and excluded from the later comparisons. Details of MLP are presented in Algorithm~\ref{alg:MLP}.

\begin{algorithm}[h!]
\caption{Multi-layer pruning (MLP)}\label{alg:MLP}
\begin{algorithmic}[1]
\\ \textbf{Input:} A dendrogram $\mathcal{D}$, number of bootstrap samples $B$ for the Gap statistics, and reference number of clusters $K$. 
\\ Assign the root node of $\mathcal{D}$ as the current node; mark current node as active.
\\ \textbf{Repeat} Step~4-7.
\\ Split the current node sequentially according to $\mathcal{D}$; obtain increasing number of clusters.
\\ Evaluate different numbers of clusters from Step~4 using the Gap statistics. 
\begin{itemize}
\\ If the chosen number of clusters is 1, set the current node as inactive; 
\\ Otherwise, split the current node into two active leaf nodes.
\end{itemize}
\\ \textbf{Break} if either of the following applies:
\begin{itemize} 
\\ Number of leaf nodes (both active and inactive) is equal to $K$.
\\ All the leaf nodes are inactive.
\end{itemize}
\\ Assign the active leaf node with the highest height in $\mathcal{D}$ as the current node.
\\ \textbf{Output: } Number of leaf nodes (less than or equal to $K$), and the corresponding cluster labels $\mathcal{L}$.
\end{algorithmic}
\end{algorithm}

The reference number of clusters $K$ can be chosen based on subject area knowledge. If not specified, we set the default reference number of clusters to be $\max\{2, K_0\}$, where $K_0$ is set as follows: apply MLP with $K=+\infty$ to the dendrograms generated from $\mathcal{C}_i$, $i = 1, \cdots, R$, then $K_0$ is the largest output number of leaf nodes from MLP. We have seen that in practice, a reliable $K_0$ can usually be found by applying MLP to $\mathcal{C}_i$, $i = 1, 2, 3$. 

Suppose there are $M$ ($M \le R$) left over dendrograms after screening out the ones with less than $K$ clusters. Let $r_j$, $\mathcal{C}_{r_j}$ and $\mathcal{L}_{r_j}$  ($j = 1, \cdots, M$) denote their corresponding ranks, feature sets and labels (for $K$ clusters from MLP), respectively. Given $\mathcal{C}_{r_j}$ and $\mathcal{L}_{r_j}$, the degree of separation of the corresponding $K$ clusters can be evaluated by the average silhouette value $S_{r_j}$. High average silhouette values indicate well-separated clusters, and thus, are preferred. If two ranks lead to similar average silhouette values, the lower one is preferred since the feature set associated with the higher rank is more likely to contain noise variables. Therefore, among local minimums in $S_{r_j}$ ($j = 1, \cdots, M$), the one with the highest rank is the least favourite, and thus, we remove such local minimums iteratively until the left-over $S_{r_j}$ are monotonically increasing or decreasing. Given monotonically increasing average silhouette values, the smallest rank after the largest increase in average silhouette value will be selected, since smaller ranks are preferred unless the increase in average silhouette value is large. On the other hand, if the average silhouette values are decreasing as rank increases, the smallest left-over rank will be selected. Details of this iterative selection approach are presented in Algorithm~\ref{alg:iter}.

\begin{algorithm}[h!]
\caption{Iterative selection of rank}\label{alg:iter}
\begin{algorithmic}[1]
\\ \textbf{Input:} $\bm{X}_{n \times p}$, $\mathcal{C}_{r_j}$ and $\mathcal{L}_{r_j}$, $j = 1, \cdots, M$.
\\ \textbf{For} $j = 1, \cdots, M$; $S_{r_j} :=$ the average silhouette value calculated from $\mathcal{C}_{r_j}$, $\mathcal{L}_{r_j}$ and $\bm{X}$.
\\ \textbf{Repeat} Step~4-5.
\\ Among the local minimums in $S_{r_j}$ ($j = 1, \cdots, M$), remove the one with the highest rank.
\\ \textbf{Break} if the left-over $S_{r_j}$, as $r_j$ increases, are monotonically:
\begin{itemize}
\item increasing: $r :=$ the smallest rank $r_j$ after the biggest increase in the left-over $S_{r_j}$.
\item decreasing: $r :=$ the smallest left-over rank $r_j$.
\end{itemize}
\\ \textbf{Output:} The chosen rank $r$. 
\end{algorithmic}
\end{algorithm}

Once the chosen rank $r$ is obtained from Algorithm~\ref{alg:iter}, MrSHC generates a dendrogram (with any linkage of choice) based on the features in $\mathcal{C}_r$.

\begin{figure}[htbp]
\begin{center}
  \includegraphics[scale = 0.25]{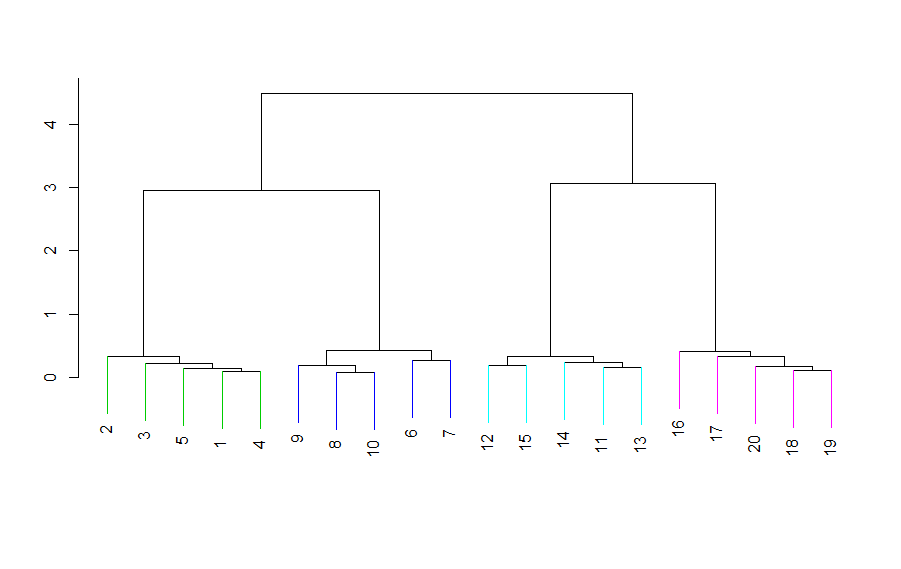}
  \end{center}
\caption{Dendrogram generated from MrSHC with known $q$}
  \label{exampleIrevisit}
\end{figure}

We revisit the example in Section~\ref{sec:limitation}. Suppose $q = 4$ is known, and we apply MrSHC with default reference number of clusters $K=2$. The resulting $C_r$ with its corresponding rank $r=2$ contains the four true clustering features: \textcolor[rgb]{0.98,0.00,0.00}{\textbf{$V_1$}}, \textcolor[rgb]{0.98,0.00,0.00}{\textbf{$V_2$}}, \textcolor[rgb]{0.98,0.00,0.00}{\textbf{$V_3$}} and \textcolor[rgb]{0.98,0.00,0.00}{\textbf{$V_4$}}. The resulting dendrogram is presented in Figure~\ref{exampleIrevisit}. The four clusters are separated correctly.

\subsection{Case 3: Unknown $q$ and $r$} \label{sec:+q+r}
Suppose $q$ and $r$ are both unknown. MrSHC considers a list of target numbers of chosen features $q_t, t = 1, \cdots, T$. For each of the candidate $q_t$, MrSHC chooses its corresponding feature set $\mathcal{C}_{q_t}$ ($|\mathcal{C}_{q_t}| = q_t$) and average silhouette value $S_{q_t}$ as described in Section~\ref{sec:+q-r}. Higher silhouette values are preferred, and at the mean time, smaller target numbers of features are preferred for the sake of interpretation and exclusion of noise features. Therefore, MrSHC uses a similar iterative approach as in Algorithm~\ref{alg:iter} to choose $q$ among $q_t$ ($t = 1, \cdots, T$). Details of this iterative approach are presented in Algorithm~\ref{alg:iter2}.

 \begin{algorithm}[h!]
\caption{Iterative selection of number of features}\label{alg:iter2}
\begin{algorithmic}[1]
\\ \textbf{Input:} $q_t$ and $S_{q_t}$, $t = 1, \cdots, T$.
\\ \textbf{Repeat} Step~3-4.
\\ Among the local minimums in $S_{q_t}$ ($t = 1, \cdots, T$), remove the one with the highest $q_t$.
\\ \textbf{Break} if the left-over $S_{q_t}$, as $q_t$ increases, are monotonically:
\begin{itemize}
\item increasing: $q :=$ the smallest $q_t$ after the biggest increase in the left-over $S_{q_t}$.
\item decreasing: $q :=$ the smallest left-over rank $q_t$.
\end{itemize}
\\ \textbf{Output:} The chosen target number of chosen features $q$. 
\end{algorithmic}
\end{algorithm}

Once the chosen $q$ is obtained from Algorithm~\ref{alg:iter2}, MrSHC generates a dendrogram (with any linkage of choice) based on the features in its corresponding $\mathcal{C}_q$.

\begin{figure}[htbp]
\begin{center}
  \includegraphics[scale = 0.18]{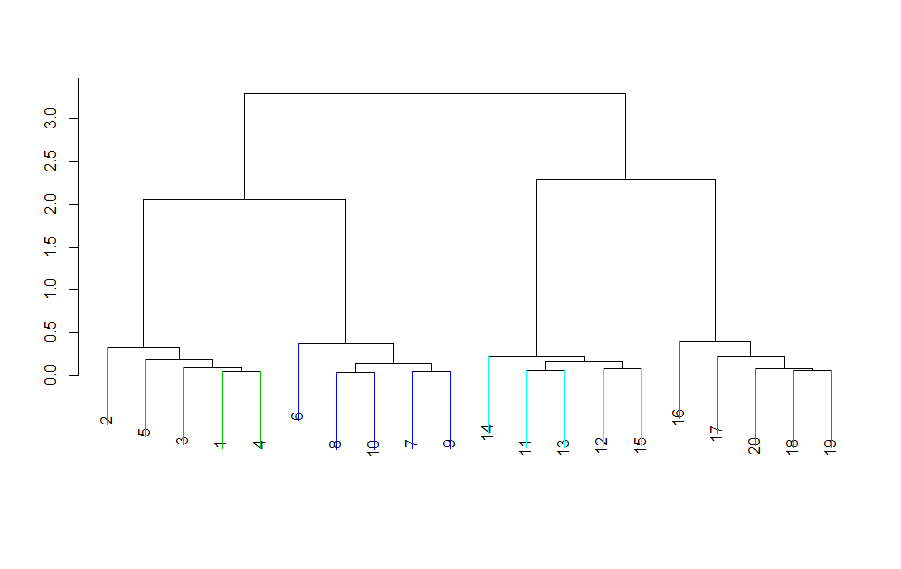}
  \end{center}
  \caption{Dendrogram generated from MrSHC with unknown $q$}
  \label{exampleIrevisit2}
\end{figure}

Again, we revisit the example in Section~\ref{sec:limitation}. Suppose $q$ and $r$ are unknown, and we apply MrSHC with default reference number of clusters $K=2$ and the list of $q_t$ \{$2, 3, \cdots, 8$\}. MrSHC suggests $q = 3$ and its corresponding feature set \{\textcolor[rgb]{0.98,0.00,0.00}{\textbf{$V1$}}, \textcolor[rgb]{0.98,0.00,0.00}{\textbf{$V3$}}, \textcolor[rgb]{0.98,0.00,0.00}{\textbf{$V4$}}\}. Although $q = 3$ is smaller than the true value $4$, the four clusters can still be separated correctly (see Figure~\ref{exampleIrevisit2}). 

\section{Simulation study}
We conduct simulation studies to compare the quality of dendrograms and the accuracy of feature selection of the following methods: HC, SHC (with known/unknown $q$) and MrSHC (with known/unknown $q$). We show the results for all the methods with complete linkage. Similar results (not shown here) are obtained with other linkages. 

\subsection{Simulation I}
We generate data sets $\mathbf X$ with $n=60$ observations and $p=500$ features as follows. The observations are generated from three main underlying clusters C1, C2 and C3. 




To be more specific, the clusters are determined by $q = 50$ features as follows:
\[
	X_{ij} = \begin{cases}
			\mu_i + \epsilon_{ij} & j=1,...,50\\
			\epsilon_{ij} & j=51,...,500
			\end{cases}
\]
where $\epsilon_{ij}\sim_{i.i.d} N(0,1)$ and 
\[
	\mu_{i} = \begin{cases}
			0 & i=1,\cdots, 20~ (i \in C1)\\
		        \mu & i = 21,\cdots, 40~ (i \in C2)\\
			-\mu & i = 41,\cdots, 60~ (i \in C3)\\
			\end{cases}
\]
We show the results for $\mu=1$. Similar conclusions are obtained for other choices of $\mu$, say $0.8$.

We generate $100$ data sets and apply HC, SHC (with known/unknown $q$) and MrSHC (with known/unknown $q$) to each. 

The quality of the resulting dendrograms is evaluated as follows. The dendrograms are cut at a level to obtain three clusters. 
The classification error rate (CER) is then used to assess clustering accuracy  \citep[see][]{chipman2006hybrid}. We do this by comparing the resulting labels from three clusters against the underlying true labels (C1, C2, C3). Given two cluster partitions, the CER is the proportion of pairs of observations that are together in one partition and apart in the other. The formulas for the CER can be found in \citet{chipman2006hybrid}.  

The accuracy of feature selection is evaluated by the recall rate (RR). Let $\mathcal J$ be the set of indices corresponding to all the clustering features, and $|\mathcal J| = q$. The recall rate (RR) is calculated as follows:
\[
	RR(\mathcal J) = \frac{\sum_{j \in \mathcal J} I(\widehat w_j \neq 0) }{q},
\]
where $I(\cdot)$ is an indicator function.


Table~\ref{tab:MrSHC1} presents the average CER, average RR, and the corresponding average $q$. For SHC and MrSHC, the unknown $q$ is chosen automatically. HC gives the highest CER, while SHC achieves better accuracy due to the sparseness. MrSHC achieves the best accuracy among the three methods. When $q = 50$ is known, SHC and MrSHC give very similar average RR. When $q$ is unknown, both methods give almost perfect RR, while MrSHC selects less features on average.
\begin{table}[h]
\centering
\small
\begin{tabular}{c|cc|ccc|ccc}
  \hline
  & \multicolumn{2}{c|}{HC} & \multicolumn{3}{c|}{SHC} & \multicolumn{3}{c}{MrSHC} \\
  \cline{2-9}
  & CER & $q$ & CER & RR & $q$ & CER & RR & $q$\\
\hline
Known ($q = 50$) & 0.202 & 500 & 0.047 & 0.926 & 50 & 0.009  & 0.995 & 50 \\ 
\hline
Unknown $q$ & 0.202 & 500 & 0.127 & 0.997 & 30.6 & 0.064 & 1.000 & 19.7 \\ 
\hline
\end{tabular}
\caption{Average CER, RR, and $q$ for different clustering methods.}\label{tab:MrSHC1}
\end{table}

\subsection{Simulation II}
We generate data sets $\mathbf X$ with $n=80$ observations and $p=500$ features, i.e.\ $\bm{X}_{n\times p}=(\bm{x}_1, \bm{x}_2, \cdots, \bm{x}_{n})^T$, where $\bm{x}_i=(x_{i,1}, x_{i,2}, \cdots, x_{i,p})^T$, $1\le i \le n$. The observations are generated from four main underlying clusters C1, C2, C3 and C4.  Let $Y_i$, ($i = 1, \cdots, n$) denote the cluster memberships. Then $x_{ij}$ ($i = 1, \cdots, n$) is generated from $N(\mu_j(Y_i), 0.1)$ for $j = 1, \cdots, 50$, and $N(\mu_j(Y_i), 1)$ for $j = 51, \cdots, p$.

A sketch of $\bm\mu_j(Y_i)$ is presented in the table below:
\begin{table}[h]
\centering
\begin{tabular}{c|cccccc|c}
  \hline
  $Y_i$ & $\mu_1(Y_i)$ & $\cdots$ & $\mu_{25}(Y_i)$ &$\mu_{26}(Y_i)$ & $\cdots$ & $\mu_{50}(Y_i)$ & $\mu_{51}(Y_i) \cdots \mu_{p}(Y_i)$\\
    \hline
  1 & $\mu$ & $\cdots$ & $\mu$ & $\mu$ & $\cdots$ & $\mu$ & $0 ~~~ \cdots ~~~ 0$\\
  2 & $-1.5\mu$  & $\cdots$ & $-1.5\mu$ & 0 & $\cdots$ & 0 &  $0 ~~~ \cdots ~~~ 0$\\
  3 & 0 & $\cdots$ & 0 & $-\mu$ & $\cdots$ & $-\mu$ &$0 ~~~ \cdots ~~~ 0$ \\
  4 & 0 & $\cdots$ & 0 & 0 & $\cdots$ & 0 &  $0 ~~~ \cdots ~~~ 0$\\
  \hline 
\end{tabular}
\end{table}

We show the results for $\mu=1$. Similar conclusions are obtained for other choices of $\mu$. 

Again, we generate $100$ data sets and apply HC, SHC (with known/unknown $q$) and MrSHC (with known/unknown $q$) to each. 
Table~\ref{tab:MrSHC2} presents the average CER, average RR, and the corresponding average $q$. When $q = 50$ is known, HC and MrSHC produce the highest and lowest average CER, respectively. MrSHC produces more accurate feature selection and clustering than SHC due to the smaller average CER and larger average RR. When $q$ is unknown, SHC produces the highest average CER with on average $19.5$ chosen features, while MrSHC achieves the smallest average CER with on average $31.5$ chosen features.
\begin{table}[h]
\centering
\small
\begin{tabular}{c|cc|ccc|ccc}
  \hline
  & \multicolumn{2}{c|}{HC} & \multicolumn{3}{c|}{SHC} & \multicolumn{3}{c}{MrSHC} \\
  \cline{2-9}
  & CER & $q$ & CER & RR & $q$ & CER & RR & $q$\\
\hline
Known ($q = 50$) & 0.172 & 500 & 0.129 & 0.875 & 50 & 0.041 & 0.977 & 50 \\ 
\hline
Unknown $q$ & 0.172 & 500 & 0.202 & 1.000 & 19.5 & 0.057 & 0.992 & 31.5 \\ 
\hline
\end{tabular}
\caption{Average CER, RR, and $q$ for different clustering methods.}\label{tab:MrSHC2}
\end{table}

\subsection{Computational times and complexity}
We investigate the computational times of MrSHC and SHC. 
The \texttt{HierarchicalSparseCluster} function from the R-package \texttt{sparcl} is used to conduct SHC. MrSHC is implemented in \texttt{R}, where the \texttt{SPC} function from the R-package \texttt{PMA} is used to conduct the sparse PCA algorithm. We use default input parameters in MrSHC: the number of bootstrap samples $B = 50$, maximum rank $R = 5$, and default selection of the reference number of clusters $K$. The average computing times for Simulation I \& II are presented in Table~\ref{tab:time}. 

\begin{table}[h]
\centering
\small
\begin{tabular}{c|cc|cc|cc}
  \hline
  & \multicolumn{2}{c|}{HC} & \multicolumn{2}{c|}{SHC} & \multicolumn{2}{c}{MrSHC} \\
  \cline{1-7}
 Simulation & I & II & I & II & I & II\\
\hline
Known ($q = 50$) & 0.006 & 0.012 & 0.746 & 1.075 & 7.754 & 11.276 \\
\hline
Unknown $q$ & 0.006 & 0.012 & 10.496 & 15.518 & 73.254 & 103.752\\
\hline
\end{tabular}
\caption{Average computing times (in seconds) for different clustering methods.}\label{tab:time}
\end{table}


When $n$ and $p$ are relatively small, SHC takes less time to compute. However, since the framework of MrSHC is embarrassingly parallel, parallel computing functions such as \texttt{mcapply} from the R-package \texttt{multicore} can be easily used to speed up the computation. Moreover, when $n$ and $p$ get larger, MrSHC will become less time consuming (observed with $n = 320$ and $p = 2000$, results not shown). This is because the computational complexities of MrSHC and SHC are $O(n^3 q B + n p)$ and $O(n^3 q + n^2 p)$, respectively, and as a result, SHC will become more computationally demanding due to the $n^2 p$ term as $n$ and $p$ increase and $p >> n$.


\section{Application to microarray data sets}
Three microarray data sets \citep{alizadeh2000distinct, perou2000molecular, van2002gene} are considered. We apply HC, SHC, MrSHC, and HC using features with the highest marginal variance (HC-HMV) to each data set. Dendrograms are created using the complete linkage. 
The default reference number of clusters is used in MrSHC, and the default list of candidate $q$ (for both SHC and MrSHC) is $\{10, 20, 40, \cdots, 100, 120, 140, \cdots, 200, 250, 300, \cdots. 500\}$.

\subsubsection{Lymphoma data set in \citet{dettling2004bagboosting}}
The data set is provided by \citet{dettling2004bagboosting}. It contains 4026 gene expression levels (features) for 62 samples (observations). Three types of most prevalent adult lymphoid malignancies were studied: 42 cases of diffuse large B-cell lymphoma (DLBCL), 9 samples of follicular lymphoma (FL), and 11 observations of B-cell chronic lymphocytic leukemia (CLL). A specialized cDNA microarray was used to measure the gene expression levels. 
Following the pre-processing steps in \citet{dudoit2002comparison}, the data set is pre-processed by first setting a thresholding window $[100, 16000]$ and then excluding genes with $\max / \min \le 5$ or $(\max- \min) \le 500$. A logarithmic transformation and standardization are applied. Finally, a simple 5 nearest neighbor algorithm is employed to impute the missing values.

The dendrograms generated from HC, SHC, MrSHC and HC-HMV are shown in Figure~\ref{fig:lymphoma}. Colors are used to indicate the three tumor types. HC only misclassifies two red samples, while SHC gives mixed clusters with the automatically chosen $q = 20$. MrSHC chooses $q = 140$ and uses rank $r = 2$, with only two blue samples misclassified into the green cluster (notice that the blue and green clusters are closer to each other). HC-HMV with $q = 140$ mixes the blue and green clusters. Therefore, MrSHC achieves better clustering accuracy with a better chosen $q = 140$ features using rank $r = 2$.

\begin{figure}[htbp]
\begin{center}
  \includegraphics[scale = 0.12]{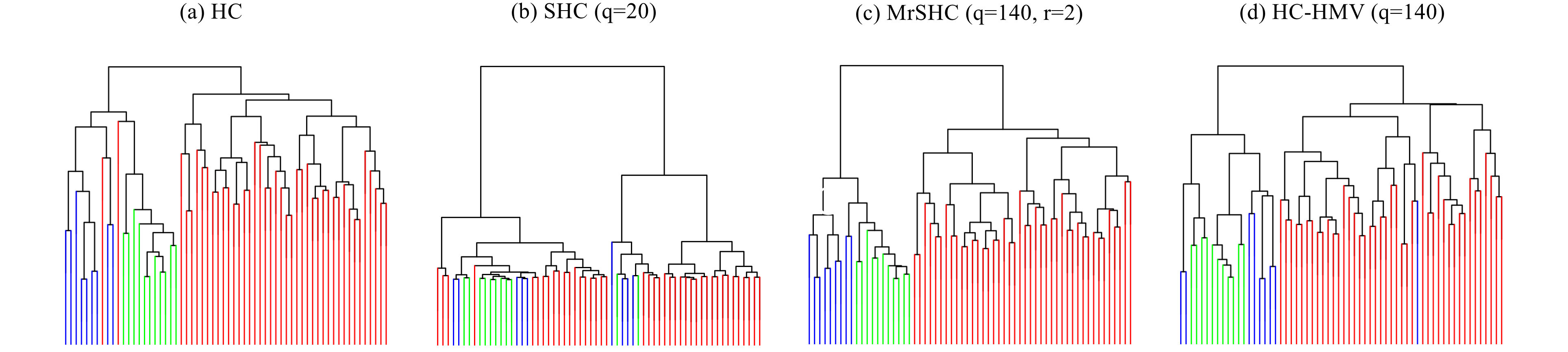}\\
  \end{center}
  \caption{Dendrograms generated by HC, SHC, MrSHC and HC-HMV for \citet{dettling2004bagboosting} data ($n=62$, $p=4026$)}
  \label{fig:lymphoma}
\end{figure}

We further investigate the effect of the rank selection by the dendrograms in Figure~\ref{???}. Figure~\ref{???}(a) shows the dendrogram generated from MrSHC with $q = 140$, but $r = 1$. Figure~\ref{???}(b) shows the dendrogram generated from SHC with $q = 140$. Both dendrograms suggest mixed clusters. This confirms that the rank selection can be crucial for better feature selection and clustering accuracy. 

\begin{figure}[htbp]
\begin{center}
  \includegraphics[scale = 0.15]{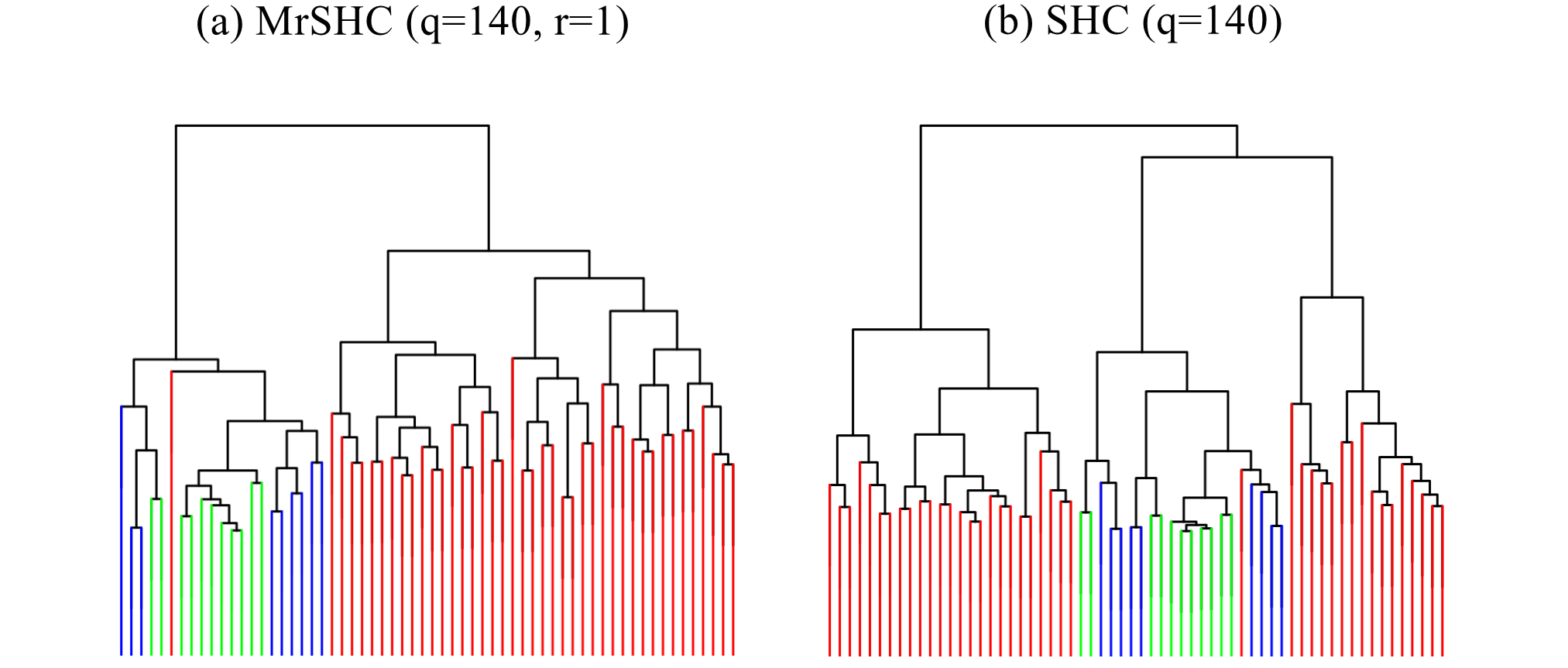}\\
  \end{center}
  \caption{Dendrograms generated by MrSHC (rank $1$) and SHC with $q=140$ for \citet{dettling2004bagboosting} data ($n=62$, $p=4026$)}
  \label{fig:lymphoma}
\end{figure}

\subsubsection{Breast cancer data set in \citet{perou2000molecular}}

The data set was first published in \citet{perou2000molecular} and later analyzed in \citet{witten2010framework}. 
 It contains 1753 gene expression levels (features) for 62 samples (observations)  to profile  surgical specimens of human breast tumors. 
 \citet{perou2000molecular} categorized the 62 samples into four groups (clusters): basal-like, Erb-B2, normal breast-like, and ER+.  \citet{perou2000molecular} suggested  that the four underlying clusters could be  explained by only 496 of the 1753 features, which was confirmed by \citet{witten2010framework}. Two misclassified samples were identified by \citet{witten2010framework}. The data set was pre-processed before being published. As such, there are no outliers in the data set.

\begin{figure}[h]
\begin{center}
  \includegraphics[scale = 0.12]{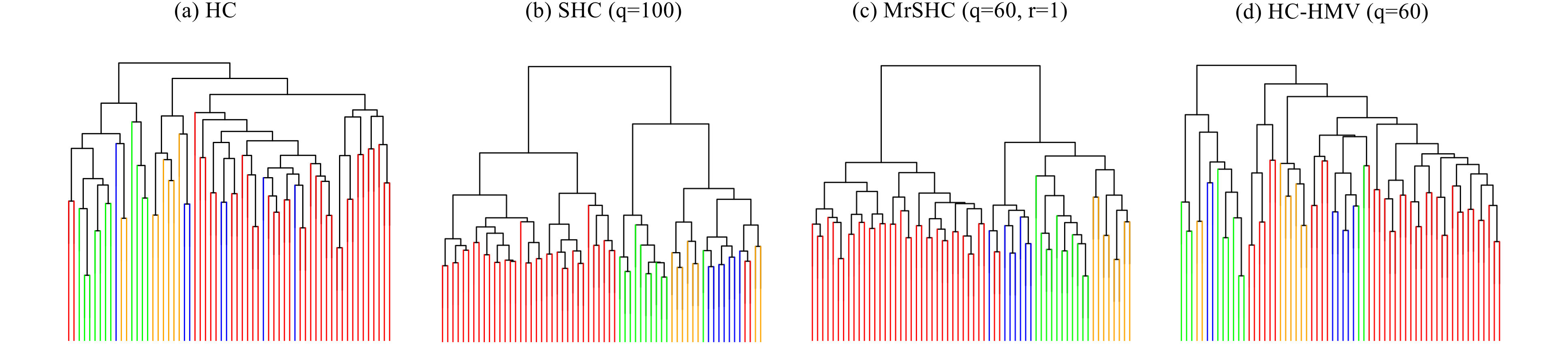}
  \end{center}
\caption{Dendrograms generated by HC, SHC, MrSHC and HC-HMV for \citet{perou2000molecular} data ($n=62$, $p=1753$)}\label{fig:witten}
\end{figure}

Figure~\ref{fig:witten} shows the dendrograms generated from HC, SHC, MrSHC and HC-HMV. Colors are used to indicate the suggested four tumor groups. HC gives mixed clusters. SHC achieves better clustering --  5 misclassified samples, with the automatically chosen $q = 100$ (similar results -- $q = 93$ features were automatically chosen and 5 samples are misclassfied -- were obtained in \citet{witten2010framework}). MrSHC misclassifies only 2 samples by using the automatically selected $q = 60$ and rank $r = 1$. HC-HMV with $q = 60$ gives mixed clusters. Although MrSHC chooses $r = 1$ over higher ranks, it still provides better feature selection and more accurate clustering comparing to the other three methods.

\subsubsection{Breast cancer data set in \citet{van2002gene}}
The data set was presented and analyzed in \citet{van2002gene}. It consists of 4751 gene expression levels for 77 primary breast tumor samples. 
A supervised classification technique was used in \citet{van2002gene}, revealing that only a subset of 70 out of the 4751 genes may help discriminating patients that have developed distant metastasis within five years. 

\begin{figure}[h]
\begin{center}
  \includegraphics[scale = 0.16]{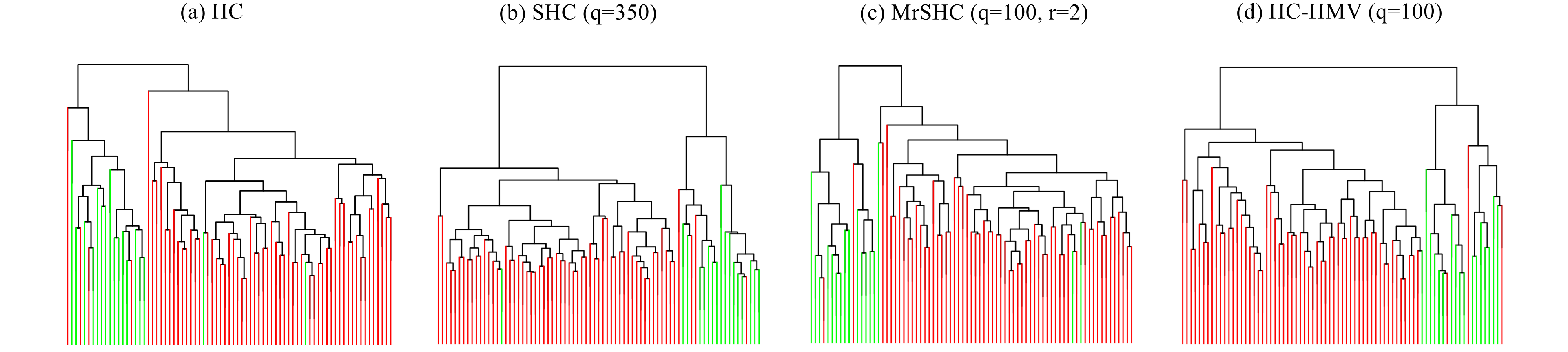}
  \end{center}
\caption{Dendrograms generated by HC, SHC, MrSHC and HC-HMV for \citet{van2002gene}'s data. ($n=77$, $p=4751$)}\label{fig:yumi}
\end{figure}

Figure~\ref{fig:yumi} shows the resulting dendrograms generated from HC, SHC, MrSHC and HC-HMV. HC misclassifies 6 samples. SHC achieves slightly better accuracy -- 5 misclassified samples, with the automatically chosen $q = 350$. MrSHC with $r = 2$ achieves the same accuracy with less ($q = 100$) features. HC-HMV with $q = 100$ features gives similar accuracy as MrSHC and SHC.

\section{Conclusion}
In this paper, we propose the multi-rank sparse hierarchical clustering (MrSHC), which automatically selects clustering features with higher rank considerations and produces the corresponding sparse hierarchical clustering. As demonstrated in simulation studies and real data examples, MrSHC gives superior feature selection and clustering performance comparing with the classical hierarchical clustering and the sparse hierarchical clustering proposed by \citet{witten2010framework}. For future research, we would like to endow MrSHC with the capability of dealing with data contamination: missing data and outliers. 


\newpage
\section*{References}
\bibliographystyle{kbib}
\bibliography{myref}

\end{document}